\pdfoutput=1

\documentclass[11pt]{article}

\usepackage{EMNLP2022}

\usepackage{times}
\usepackage{latexsym}

\usepackage[T1]{fontenc}

\usepackage[utf8]{inputenc}

\usepackage{microtype}

\usepackage{inconsolata}

\usepackage{times}
\usepackage{latexsym}

\usepackage{color,soul} 
\usepackage{graphicx}
\usepackage{colortbl}
\usepackage{amsmath}
\usepackage{float}

\usepackage{microtype}
\usepackage{microtype}
\usepackage{multirow}
\usepackage{multicol}
\usepackage{soul}
\usepackage{multirow}
\usepackage{makecell}
\usepackage[utf8]{inputenc}
\usepackage{microtype}
\usepackage{graphicx}
\usepackage{soul}
\usepackage{float}
\usepackage{comment}
\usepackage{adjustbox}
\usepackage{booktabs}
\usepackage{amsmath}
\usepackage{cleveref}
\usepackage{mathabx, nicefrac}
\usepackage[shortlabels]{enumitem}
\usepackage{subcaption}

\usepackage{multicol}
\usepackage{caption}
\usepackage{subcaption}
\usepackage{color,soul}
\usepackage{microtype}
\usepackage{dblfloatfix}
\title{`Hardness' of Samples Need to be Quantified for a Reliable Evaluation System: Exploring Potential Opportunities with a New Task}

\author{Swaroop Mishra $\;$ Anjana Arunkumar $\;$ Chris Bryan $\;$ Chitta Baral
\\\\
 Arizona State University }

\date{}

\begin{document}
\maketitle

\begin{abstract}
Evaluation of models on benchmarks is unreliable without knowing the degree of sample hardness; this subsequently overestimates the capability of AI systems and limits their adoption in real world applications. We propose a \textit{Data Scoring task} that requires assignment of each unannotated sample in a benchmark a score between 0 to 1, where 0 signifies easy and 1 signifies hard. Use of unannotated samples in our task design is inspired from humans who can determine a question difficulty without knowing its correct answer. This also rules out the use of methods involving model based supervision (since they require sample annotations to get trained), eliminating potential biases associated with models in deciding sample difficulty. We propose a method based on Semantic Textual Similarity (STS) for this task; we validate our method by showing that existing models are more accurate with respect to the easier sample-chunks than with respect to the harder sample-chunks. Finally we demonstrate five novel applications. 

\end{abstract}

\section{Introduction}




Empirical justification of model superiority without theoretical proof can be unreliable. Suppose model $M_1$ has higher accuracy than model $M_2$ on a dataset $D_1$. In order to confirm if we should choose $M_1$ over $M_2$, we can train and evaluate both models on $D_2$, $D_3$ and $D_4$. Even if $M_1$ has higher accuracy than $M_2$ on all the three datasets, we cannot say that $M_1$ is always better; we may then evaluate both models on Out of Distribution (OOD) \cite{quionero2009dataset} datasets (usually done zero-shot \cite{bras2020adversarial}) $D_5$ and $D_6$ to find their generalization performance. Suppose that $M_1$ still performs better than $M_2$ in $D_5$ and $D_6$. Is that enough to justify the superiority of $M_1$, or do we need further experimentation on a wider range of datasets? Often, $M_1$ will not be a clear winner in all six datasets, making this evaluation even harder.

Justifying superiority as above requires analysis of `hardness' of questions that models are answering. For instance, $D_1$-$D_4$ may contain `easy' questions on which $M_1$ excels and `hard' questions on which it fails. It is also possible that $M_2$ answers more `hard' questions than $M_1$, while the overall count of correctly answered questions for $M_2$ is lower, so $M_1$ has better accuracy. In that case, zeroshot OOD performance can help identify the winner, assuming that OOD samples are harder samples that diverge from the training set. 

Consequently, a clear definition of the `hardness' of data items is useful in better analyzing empirical results.  It will help in the design of a futuristic testbed consisting of a hierarchy of question sets with increasing levels of difficulty, similar to hierarchical testbeds used in software engineering \cite{barreiros2011systematic} and in competitive examinations such as GRE. This will further allow us to quantify the weightage assigned to each question based on its hardness and assign weighted score to models based on their performance across various datasets, instead of the average sample performance typically calculated within a dataset and average dataset performance calculated in a benchmark (e.g. GLUE score~\cite{wang2018glue}). Additionally,  defining OOD in terms of `hardness' allows for the representation of such samples as extensions of IID (independent and identically distributed)-- `very hard' samples-- and also for the implicit identification of OOD.

\citet{swayamdipta2020dataset} uses annotated datasets to find sample hardness with a model-dependent method; however knowledge of correct answers is not a prerequisite for humans to decide on  question `hardness' (and subsequent distribution shift). Also, a model based and annotation dependent method can contain artifact as a sample which is easy for this model may be difficult for another model. To the best of our knowledge, we do not have a measure to find hardness of unannotated data.
This motivates us to propose a \textit{Data Scoring Task} and a STS based method that assigns each unannotated sample a score between 0 (`easy') and 1 (`hard'), and explains distribution shift, thus quantifying the degree of OOD characteristics. Here, `hardness' is considered to be inversely proportional to model predictability. For instance, if question $q_1$ is a random sample from set $s_1$, and $q_2$ is a random sample from set $s_2$ such that $s_2$ is harder than $s_1$, then the probability that $q_1$ will be correctly answered by a model $M$ is higher than for $q_2$, i.e., model predictability of $q_1$ is higher than that of $q_2$.


\textbf{Contributions:}
In summary, the contribution of this work are as follows: (i) We propose a \textit{Data Scoring task} realizing the need for quantification of `hardness' of samples to build a reliable evaluation system, (ii) We propose an STS based approach that gives strong results, specially on recent transformers, (iii) We demonstrate five novel applications and opportunities associated with this task. 


\section{Our Approach:}
First, we propose an annotation-agnostic measure that quantifies a sample's relative predictability (`hardness'), and explains relative distribution shift. 

 We experiment with ten models across an IID-OOD dataset pair, using STS (Semantic Textual Similarity) of test samples with respect to the training set for our task, and find it a strong indicator of predictability in regular/zero shot OOD settings. 

We analyze several recent works \cite{hendrycks2020pretrained, bras2020adversarial, hendrycks2019benchmarking, talmor2019multiqa} involving datasets that have been paired with an OOD counterpart. Identification of OOD datasets as well as how to break ties if $M_1$ and $M_2$ excel on equal numbers of the OOD datasets used for evaluation, have remained unanswered. To address this, we divide the datasets into several hierarchies, based on STS. We can therefore reasonably identify samples within a dataset, which have higher OOD characteristic levels. This allows the same dataset to be used to evaluate OOD. STS can thus be used to draw a boundary between IID and OOD, and to control the degree of OOD characteristics in a dataset. 

Next, we formulate an equitable evaluation metric \textit{Weighting Out of Distribution Score (WOOD Score)}, that weights each test sample in proportion to its degree of OOD characteristics (`hardness') and  penalizes incorrect answers, such that `hard' samples are both `high risk' and `high gain'. This compels a model to solve `hard' questions and thus generalize in order to dominate leaderboards. 

Models that surpass human performance are often found to depend on spurious bias-- unintended correlation between input and output, \cite{bras2020adversarial}-- instead of truly learning the task \cite{gururangan2018annotation, kaushik2018much}, and thus fail to generalize on OOD data \cite{eykholt2018robust, jia2017adversarial}, leading to overestimation of AI \cite{sakaguchi2019winogrande,hendrycks2019natural}.  \textit{WOOD Score} shows a decrease in model performance, thus addressing model inflation.

Conventionally, MaxProb is used as a strong baseline \cite{hendrycks2016baseline} for misclassification and OOD detection. However, it requires data annotation/running models; thus it cannot be used in our model/annotation-agnostic task. We find that STS indicates relative distribution of MaxProb across a dataset, and can therefore be used instead of MaxProb in various tasks e.g. selective answering \cite{kamath2020selective, varshney2020s}.

Our task also allows for the selection of question subset for annotation--`hard' questions require explicit annotation-- to maintain a desired data quality level. This drastically reduces heavy resource investment in annotating and wastage~\cite{bras2020adversarial, mishra2020we} due to sample deletion (to get rid of spurious bias). This also allows for the recommendation of `hard' question creation by experts, if such questions are found to be lacking in scored datasets. This can be further extended to potentially create a hierarchical testbed, where samples from different datasets are pooled and ranked based on their hardness, leading to a new demarkation of dataset boundaries. This will ensure reliable and standardized model evaluation.

\section{Data Scoring with STS}
\label{sec:diff}
We use two movie review datasets: SST-2 \cite{socher2013recursive} and IMDB \cite{maas2011learning}, which contain succinct expert reviews and full length general reviews respectively. We utilize IMDB as the IID dataset and SST-2 as the OOD dataset, and evaluate them using ten models: Bag-of-words (BoW) model \cite{harris1954distributional}, word embedding - word2vec \cite{mikolov2013distributed} and GloVe \cite{pennington2014glove} encoded with three models -word averages \cite{wieting2015towards}, LSTM \cite{hochreiter1997long} and CNN \cite{lecun1995convolutional}, and pretrained transformer models -BERT Base and Large \cite{devlin2018bert} along with GELU \cite{hendrycks2016gaussian} and RoBERTA \cite{liu2019roberta}, following a recent work on OOD Robustness \cite{hendrycks2020pretrained}.\\
\begin{figure}[H]
    \centering
    \includegraphics[width=0.48\textwidth]{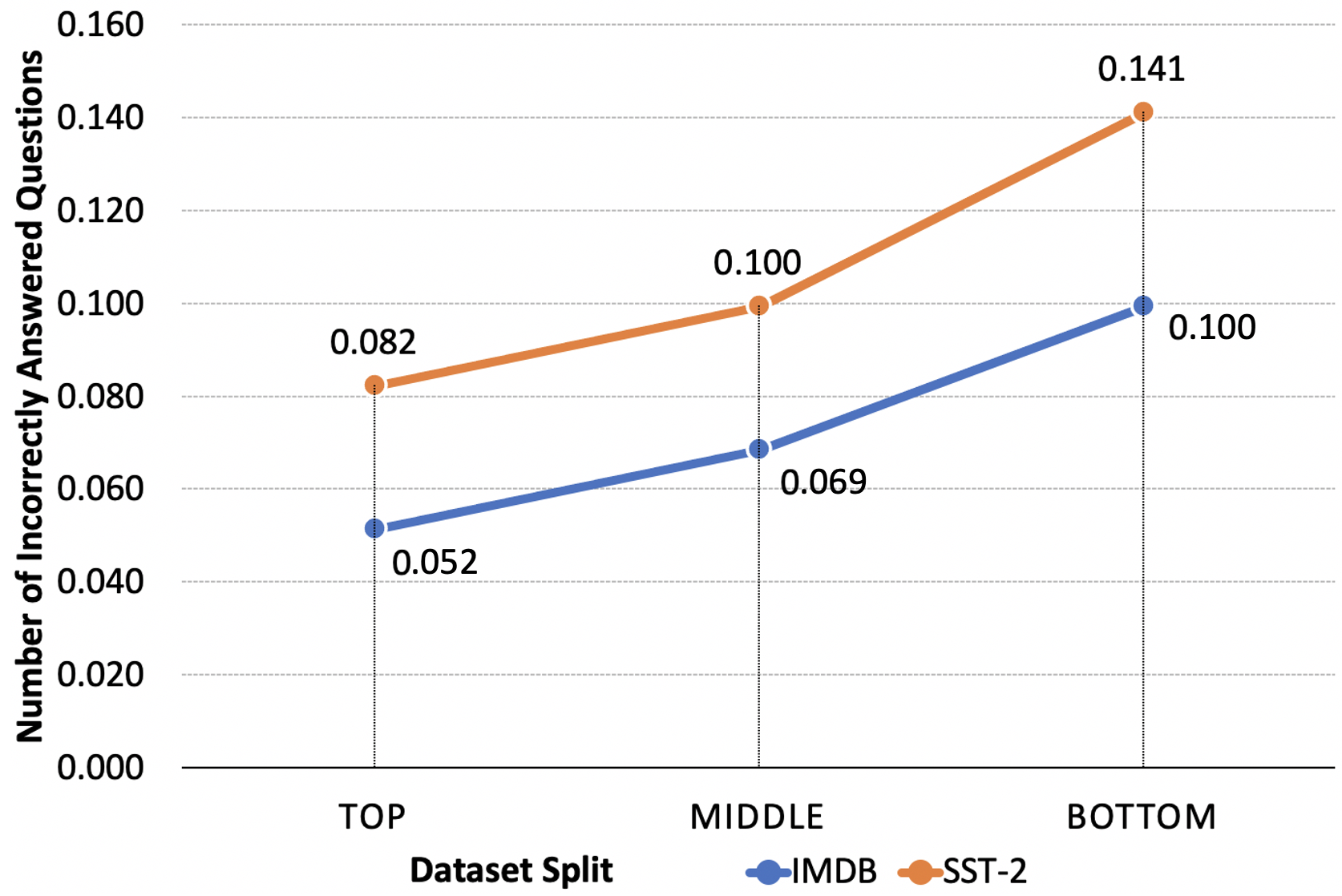}
    \caption{Percentage of incorrect classifications using BERT-Base model across test samples of SST-2 and IMDB in decreasing order of train (IMDB)-test similarity. Monotonic increase in slope is desirable.}
    \label{fig:2}
\end{figure}
\textbf{Implementation:} We use Spacy's \cite{honnibal2017spacy} BERT STS implementation to find the similarity between every pair of train-test set samples. We sort samples of the test set in a descending order, based on the average STS value with varying percentages of the SST-2 train set samples. We consider the top 1\% -- 100\% of the training data (obtained by sorting train set samples in descending order of STS against each test set sample) with nine total steps \footnote{1\%,5\%,10\%,25\%,30\%,40\%,50\%,75\%,100\%}, as similarity between the train and test sets is a task dependent hyperparameter, that trades off between inductive bias and spurious bias \cite{mishra2020dqi}. We train models on the IID data (IMDB) and evaluate on both the IID test set (IMDB) and the OOD test set (SST-2). We compare model predictions with the average STS value for each sample.\\

\begin{figure}[H]
    \centering
    \includegraphics[width=0.48\textwidth]{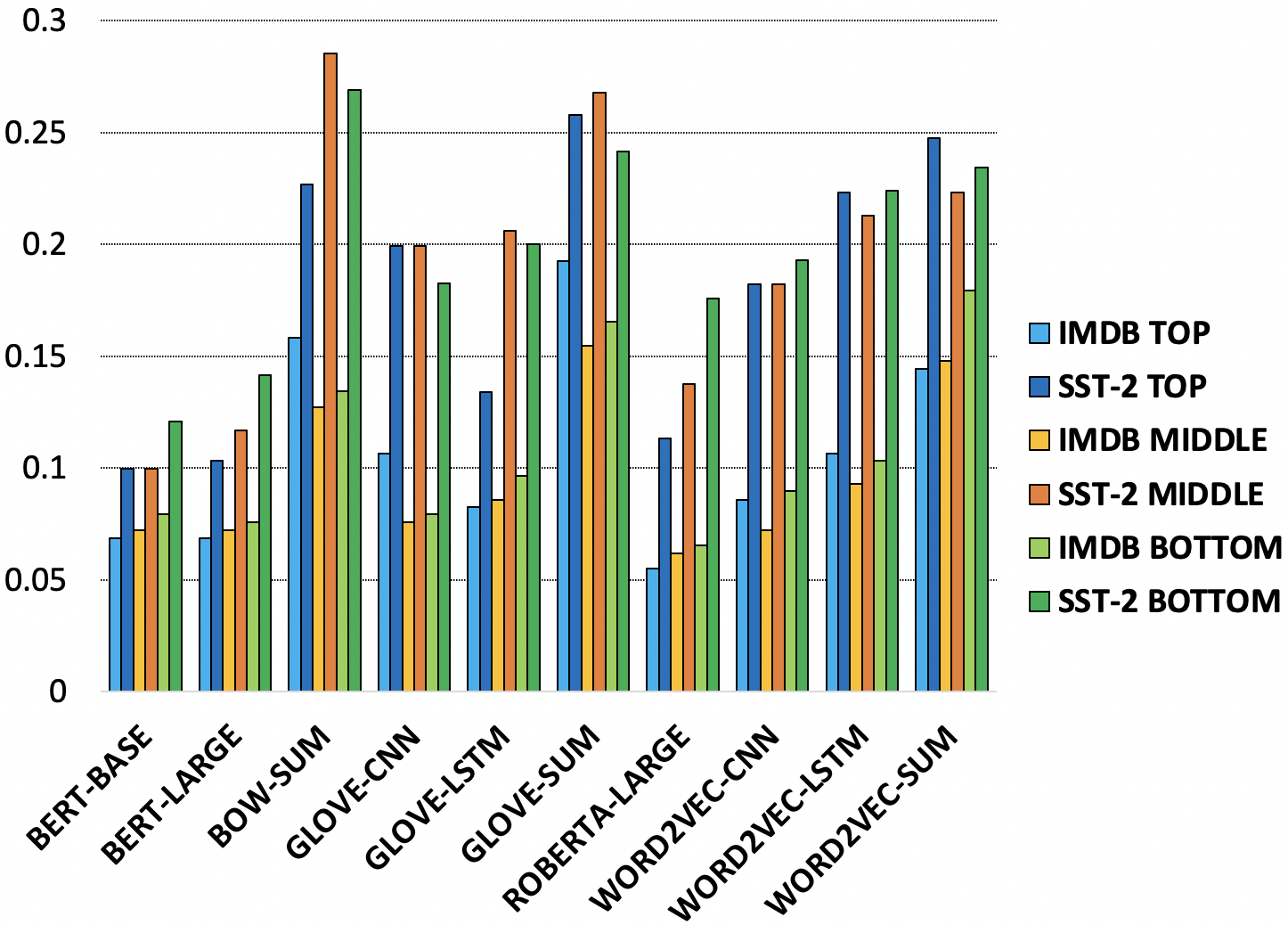}
    \caption{Percentage of incorrect classifications across test samples of SST-2 and IMDB in decreasing order of train (IMDB)-test similarity.}
    \label{fig:wrongchunks}
\end{figure}

\textbf{Results:} We find three broad patterns: (i) Sample-chunks with higher average STS (`easier samples') have fewer percentage of incorrect predictions (Figure \ref{fig:2}); in Figure \ref{fig:wrongchunks}, we show that for transformer models (in both datasets) and word2vec embedding models (in IMDB) exhibit this behavior. Our observation of STS accurately explaining predictability of transformers may indicate that Transformers are better at leveraging training data for memorization. (ii) IID sample-chunks have higher average STS than OOD  (Figure \ref{fig:comparests}); STS therefore helps in drawing a boundary between IID and OOD, and (iii) Samples with higher average STS value are classified correctly with higher confidence, and incorrectly with lower confidence (Figure \ref{fig:3}).\footnote{More details in Supplementary Material}

\section{WOOD Score for Equitable Evaluation} 
We propose \textit{equitable data evaluation} using \textit{WOOD Score}, in lieu of a conventional evaluation metric (e.g. accuracy) that weights all samples uniformly.\\
\textbf{Formalization:}
Let $X$ represent a dataset where $X_{Test}$ is the test set spanned by $i$ and $X_{Train}$ is the train set.  $E$ represents the evaluation metric (which depends on the application-- here we consider +1 for correct answers and a -1 penalty for incorrect answers). $p$ is the degree of OOD characteristics (i.e., data score) a sample has, and $S$ represents STS. $a$ allows for the control of $p$ based on $S$, $b$ is the number of train samples considered that have higher similarity values than rest of the dataset. $W_{opt}$ represents our proposed metric in generic form, and $W_{acc}$ is the proposed accuracy metric in this paper. We divide the dataset into three sample-chunks, $c_1, c_2, c_3$ having the highest, moderate, and lowest degrees of OOD characteristics respectively. 

\begin{figure*}[t]
    \includegraphics[width=2\columnwidth]{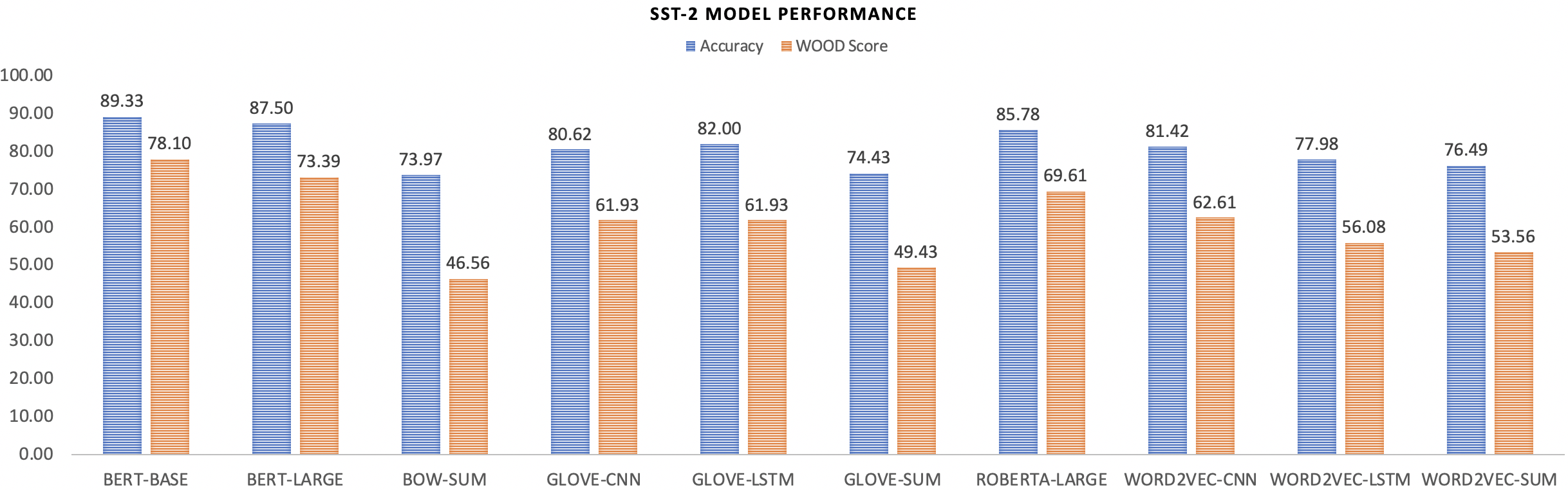}
    \includegraphics[width=2\columnwidth]{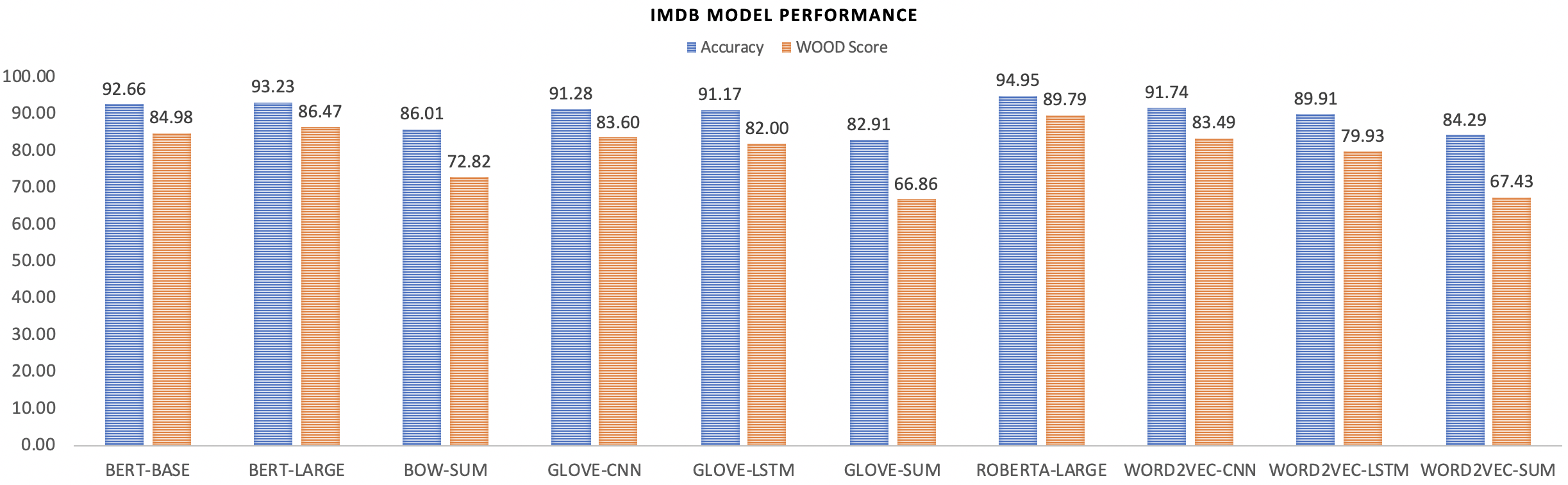}
    \caption{Accuracy and WOOD Score of SST-2 and IMDB across models. \textit{WOOD Score} is significantly lower than accuracy for both datasets, with greater decrease seen for OOD data. Ranking changes for 9/10 models in IMDB (IID) and 3/10 models in SST-2 (OOD).}
    \label{fig:4}
\vspace{-4mm}
\end{figure*}

\begin{figure}
    \centering
    \includegraphics[width=0.44\textwidth]{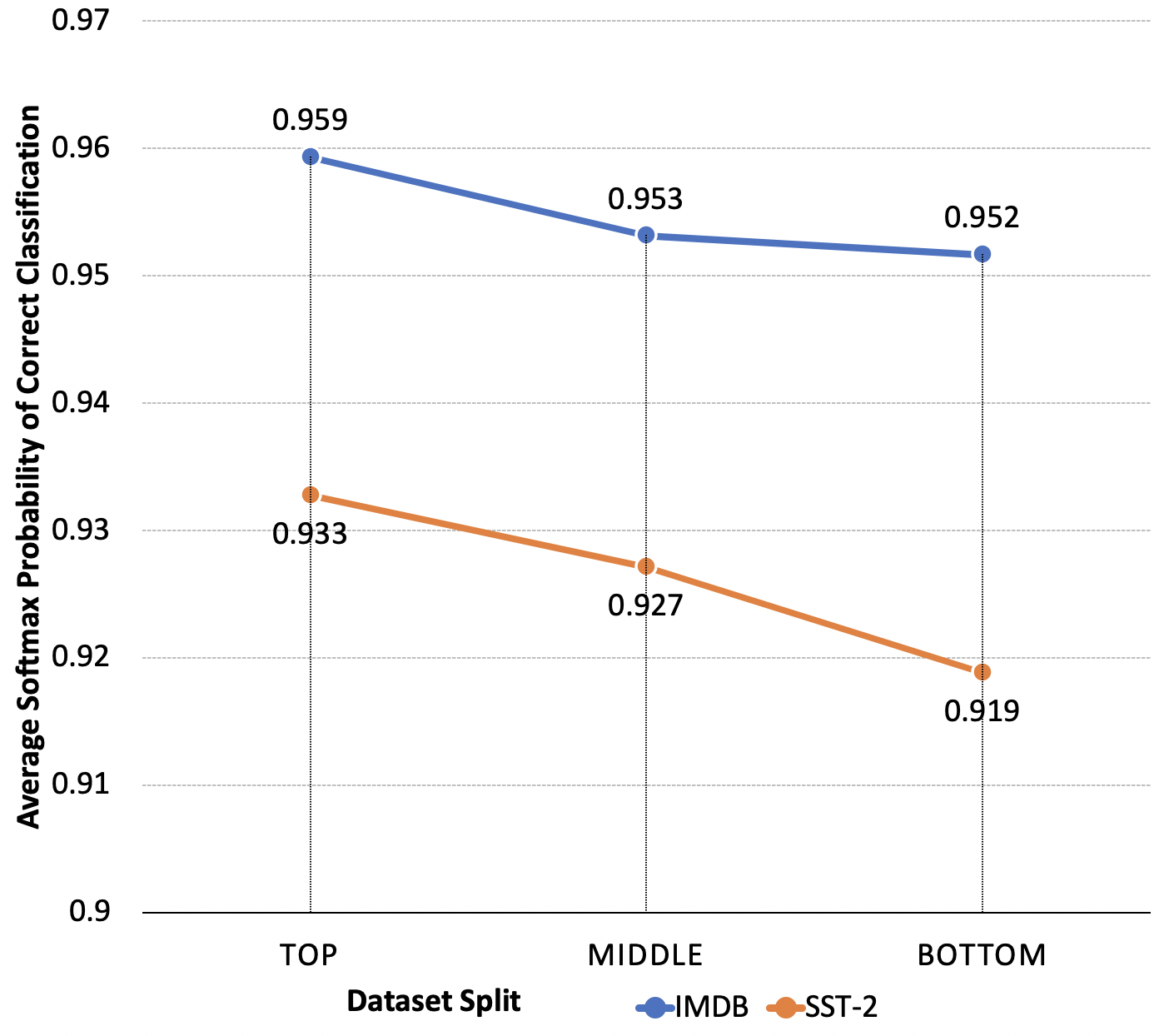}
    \caption{Average MaxProb of correct classifications using BERT-Base model across test sample-chunks of SST-2 and IMDB in decreasing order of train (IMDB)-test similarity. Monotonic slope decrease is desirable.}
    \label{fig:3}
\end{figure}
\begin{equation}
W_{opt}=\frac{\sum_{X_{Test}} E_{i}p_{i}}{\sum p_i}
\end{equation}
\begin{equation}
p=\frac{a}{\sum_{X_{Train}}{\max\limits_{b}S}}
\end{equation}
\begin{align}
W_{acc}=\frac{\sum_{X_{Test}} E_{i}p_{i}}{\sum p_i}\ \ \text{, where (based on} \max\limits_{b}S \text{)}      
\end{align}
\begin{equation*}
    p_i=
    \begin{cases}
      3 &\text{if}\ i \ \epsilon \ c_1 \\
      2 &\text{if}\ i \ \epsilon \ c_2 \\
      1 &\text{if}\ i \ \epsilon \ c_3 
    \end{cases}
\end{equation*}

\noindent\textbf{Controlling Benchmark Accuracy Using Hyperparameters:} Benchmark  accuracy can be controlled using $a$ and $b$ appropriately. $E$ controls penalties imposed for incorrect answers (e.g. safety critical applications require higher penalties).\\
Using $W_{acc}$  for both datasets across ten models has resulted in a significant reduction in accuracy, thus addressing model performance inflation (Table \ref{fig:4}, where $a$ is 1 and $b$ is taken as 0.1; however similar observations are noted for all hyperparameters). We also note that the WOOD score rankings significantly differ from that with accuracy.

\section{Discussion}
\begin{figure}
    \centering
    \includegraphics[width=0.48\textwidth]{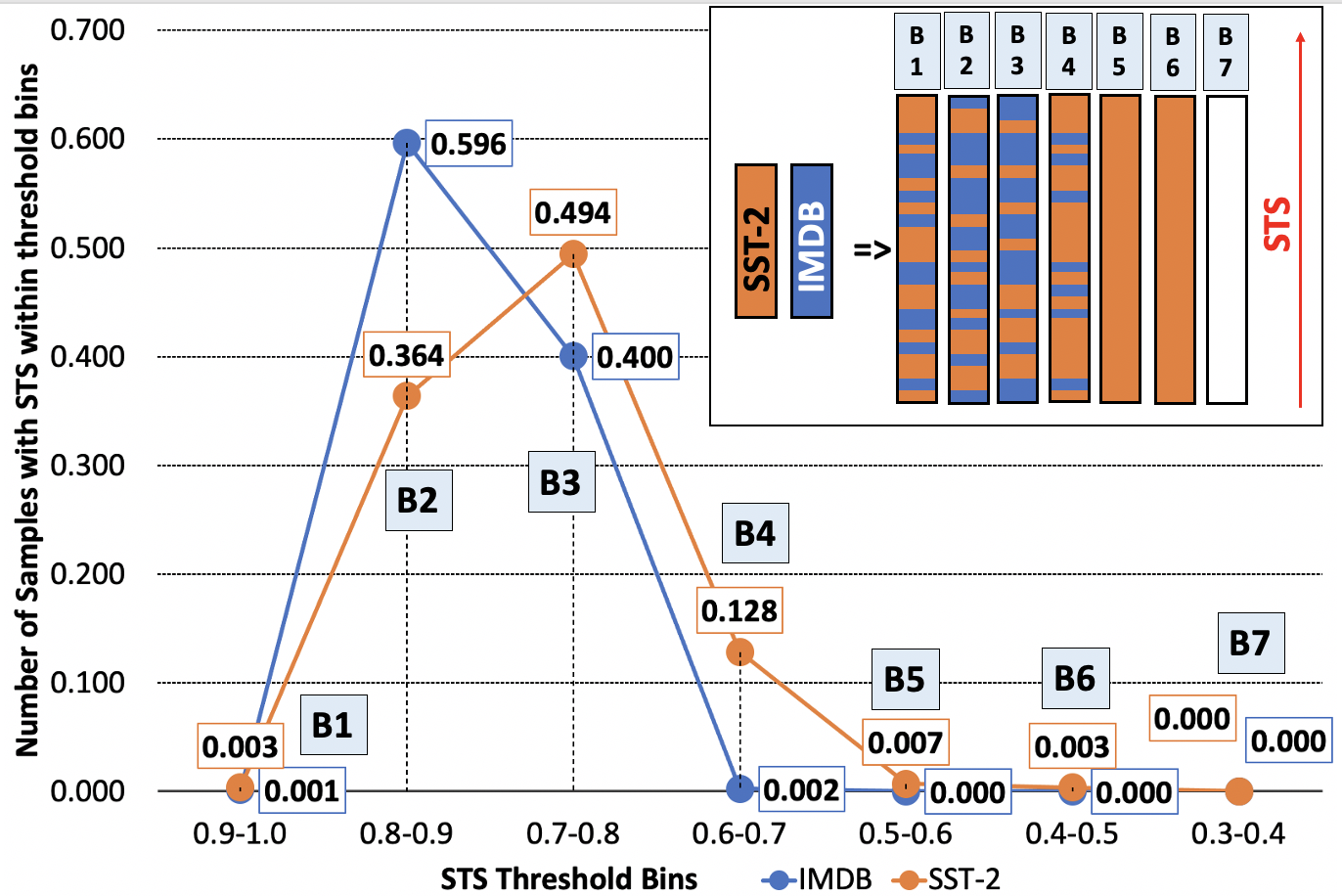}
    \caption{Percentage of samples in IMDB and SST-2 with STS value within threshold bins, with hierarchical testbed formation, from `easy' (B1) to `hard' (B7).}
    \label{fig:hard}
\end{figure}

\textbf{Data Annotation and Creation:} Figure \ref{fig:hard} shows the distribution of STS values across both datasets, and hierachical testbed creation. Assuming that these data are unannotated, authors can decide to annotate only the hard samples (e.g. below STS threshold of 0.7) as easy samples won't be really help to increase performance of a model already trained with varieties of dataset of the same task (sentiment analysis). Similarly, authors may decide to manually create hard samples (e.g. STS $<$0.5) since they are limited in both datasets. This  technique can be helpful in the learning from instruction paradigm~\cite{mishra2022cross, wei2021finetuned, sanh2021multitask,  ouyang2022training, mishra-etal-2022-reframing, parmar2022boxbart} because of the implicit setting which is low resource with annotated data.\\

\vspace{-5mm}
\section{Conclusion}

We propose a \textit{Data Scoring measure} that quantifies the hardness of each sample in an unannotated dataset based on STS, and show its applications in various domains. We further show that STS sometimes fails to appropriately indicate model predictability, demonstrating room for future research on this unexplored task. 
\section{Limitations}
\textbf{Augumenting STS:} STS may not follow monotonic behavior with model performance for certain cases, as illustrated in Figure \ref{fig:5}. Similarity across several granularities -- such as word, bigram, and trigram -- can be used to augument STS and increase the robustness of `hardness' evaluation.\\
\textbf{Strengthening `in-house' IID (acting OOD): }
We further observe that, IID data, even with STS calibration, may not represent many properties of an OOD data sample -- such as variations in writing style, topic, vocabulary , sentence length, and number of sentences. 
We recommend that dataset creators go beyond the common patterns found in a dataset, and draw patterns from other
datasets intended for the same task, while creating contrast sets \cite{gardner2020evaluating}, to address this.\\


\bibliography{anthology,custom}

\begin{thebibliography}{40}
\expandafter\ifx\csname natexlab\endcsname\relax\def\natexlab#1{#1}\fi

\bibitem[{Barreiros et~al.(2011)Barreiros, Almeida, Saraiva, and
  Soares}]{barreiros2011systematic}
Emanoel Barreiros, Adauto Almeida, Juliana Saraiva, and Sergio Soares. 2011.
\newblock A systematic mapping study on software engineering testbeds.
\newblock In \emph{2011 International Symposium on Empirical Software
  Engineering and Measurement}, pages 107--116. IEEE.

\bibitem[{Bras et~al.(2020)Bras, Swayamdipta, Bhagavatula, Zellers, Peters,
  Sabharwal, and Choi}]{bras2020adversarial}
Ronan~Le Bras, Swabha Swayamdipta, Chandra Bhagavatula, Rowan Zellers,
  Matthew~E Peters, Ashish Sabharwal, and Yejin Choi. 2020.
\newblock Adversarial filters of dataset biases.
\newblock \emph{arXiv preprint arXiv:2002.04108}.

\bibitem[{Devlin et~al.(2018)Devlin, Chang, Lee, and
  Toutanova}]{devlin2018bert}
Jacob Devlin, Ming-Wei Chang, Kenton Lee, and Kristina Toutanova. 2018.
\newblock Bert: Pre-training of deep bidirectional transformers for language
  understanding.
\newblock \emph{arXiv preprint arXiv:1810.04805}.

\bibitem[{Eykholt et~al.(2018)Eykholt, Evtimov, Fernandes, Li, Rahmati, Xiao,
  Prakash, Kohno, and Song}]{eykholt2018robust}
Kevin Eykholt, Ivan Evtimov, Earlence Fernandes, Bo~Li, Amir Rahmati, Chaowei
  Xiao, Atul Prakash, Tadayoshi Kohno, and Dawn Song. 2018.
\newblock Robust physical-world attacks on deep learning visual classification.
\newblock In \emph{Proceedings of the IEEE Conference on Computer Vision and
  Pattern Recognition}, pages 1625--1634.

\bibitem[{Gardner et~al.(2020)Gardner, Artzi, Basmova, Berant, Bogin, Chen,
  Dasigi, Dua, Elazar, Gottumukkala et~al.}]{gardner2020evaluating}
Matt Gardner, Yoav Artzi, Victoria Basmova, Jonathan Berant, Ben Bogin, Sihao
  Chen, Pradeep Dasigi, Dheeru Dua, Yanai Elazar, Ananth Gottumukkala, et~al.
  2020.
\newblock Evaluating nlp models via contrast sets.
\newblock \emph{arXiv preprint arXiv:2004.02709}.

\bibitem[{Gardner et~al.(2018)Gardner, Grus, Neumann, Tafjord, Dasigi, Liu,
  Peters, Schmitz, and Zettlemoyer}]{gardner2018allennlp}
Matt Gardner, Joel Grus, Mark Neumann, Oyvind Tafjord, Pradeep Dasigi, Nelson
  Liu, Matthew Peters, Michael Schmitz, and Luke Zettlemoyer. 2018.
\newblock Allennlp: A deep semantic natural language processing platform.
\newblock \emph{arXiv preprint arXiv:1803.07640}.

\bibitem[{Gururangan et~al.(2018)Gururangan, Swayamdipta, Levy, Schwartz,
  Bowman, and Smith}]{gururangan2018annotation}
Suchin Gururangan, Swabha Swayamdipta, Omer Levy, Roy Schwartz, Samuel~R
  Bowman, and Noah~A Smith. 2018.
\newblock Annotation artifacts in natural language inference data.
\newblock \emph{arXiv preprint arXiv:1803.02324}.

\bibitem[{Harris(1954)}]{harris1954distributional}
Zellig~S Harris. 1954.
\newblock Distributional structure.
\newblock \emph{Word}, 10(2-3):146--162.

\bibitem[{Hendrycks and Dietterich(2019)}]{hendrycks2019benchmarking}
Dan Hendrycks and Thomas Dietterich. 2019.
\newblock Benchmarking neural network robustness to common corruptions and
  perturbations.
\newblock \emph{arXiv preprint arXiv:1903.12261}.

\bibitem[{Hendrycks and Gimpel(2016{\natexlab{a}})}]{hendrycks2016baseline}
Dan Hendrycks and Kevin Gimpel. 2016{\natexlab{a}}.
\newblock A baseline for detecting misclassified and out-of-distribution
  examples in neural networks.
\newblock \emph{arXiv preprint arXiv:1610.02136}.

\bibitem[{Hendrycks and Gimpel(2016{\natexlab{b}})}]{hendrycks2016gaussian}
Dan Hendrycks and Kevin Gimpel. 2016{\natexlab{b}}.
\newblock Gaussian error linear units (gelus).
\newblock \emph{arXiv preprint arXiv:1606.08415}.

\bibitem[{Hendrycks et~al.(2020)Hendrycks, Liu, Wallace, Dziedzic, Krishnan,
  and Song}]{hendrycks2020pretrained}
Dan Hendrycks, Xiaoyuan Liu, Eric Wallace, Adam Dziedzic, Rishabh Krishnan, and
  Dawn Song. 2020.
\newblock Pretrained transformers improve out-of-distribution robustness.
\newblock \emph{arXiv preprint arXiv:2004.06100}.

\bibitem[{Hendrycks et~al.(2019)Hendrycks, Zhao, Basart, Steinhardt, and
  Song}]{hendrycks2019natural}
Dan Hendrycks, Kevin Zhao, Steven Basart, Jacob Steinhardt, and Dawn Song.
  2019.
\newblock Natural adversarial examples.
\newblock \emph{arXiv preprint arXiv:1907.07174}.

\bibitem[{Hochreiter and Schmidhuber(1997)}]{hochreiter1997long}
Sepp Hochreiter and J{\"u}rgen Schmidhuber. 1997.
\newblock Long short-term memory.
\newblock \emph{Neural computation}, 9(8):1735--1780.

\bibitem[{Honnibal and Montani(2017)}]{honnibal2017spacy}
Matthew Honnibal and Ines Montani. 2017.
\newblock spacy 2: Natural language understanding with bloom embeddings,
  convolutional neural networks and incremental parsing.
\newblock \emph{To appear}, 7(1).

\bibitem[{Jia and Liang(2017)}]{jia2017adversarial}
Robin Jia and Percy Liang. 2017.
\newblock Adversarial examples for evaluating reading comprehension systems.
\newblock \emph{arXiv preprint arXiv:1707.07328}.

\bibitem[{Kamath et~al.(2020)Kamath, Jia, and Liang}]{kamath2020selective}
Amita Kamath, Robin Jia, and Percy Liang. 2020.
\newblock Selective question answering under domain shift.
\newblock \emph{arXiv preprint arXiv:2006.09462}.

\bibitem[{Kaushik and Lipton(2018)}]{kaushik2018much}
Divyansh Kaushik and Zachary~C Lipton. 2018.
\newblock How much reading does reading comprehension require? a critical
  investigation of popular benchmarks.
\newblock \emph{arXiv preprint arXiv:1808.04926}.

\bibitem[{LeCun et~al.(1995)LeCun, Bengio et~al.}]{lecun1995convolutional}
Yann LeCun, Yoshua Bengio, et~al. 1995.
\newblock Convolutional networks for images, speech, and time series.
\newblock \emph{The handbook of brain theory and neural networks},
  3361(10):1995.

\bibitem[{Liu et~al.(2019)Liu, Ott, Goyal, Du, Joshi, Chen, Levy, Lewis,
  Zettlemoyer, and Stoyanov}]{liu2019roberta}
Yinhan Liu, Myle Ott, Naman Goyal, Jingfei Du, Mandar Joshi, Danqi Chen, Omer
  Levy, Mike Lewis, Luke Zettlemoyer, and Veselin Stoyanov. 2019.
\newblock Roberta: A robustly optimized bert pretraining approach.
\newblock \emph{arXiv preprint arXiv:1907.11692}.

\bibitem[{Maas et~al.(2011)Maas, Daly, Pham, Huang, Ng, and
  Potts}]{maas2011learning}
Andrew~L Maas, Raymond~E Daly, Peter~T Pham, Dan Huang, Andrew~Y Ng, and
  Christopher Potts. 2011.
\newblock Learning word vectors for sentiment analysis.
\newblock In \emph{Proceedings of the 49th annual meeting of the association
  for computational linguistics: Human language technologies-volume 1}, pages
  142--150. Association for Computational Linguistics.

\bibitem[{Mikolov et~al.(2013)Mikolov, Sutskever, Chen, Corrado, and
  Dean}]{mikolov2013distributed}
Tomas Mikolov, Ilya Sutskever, Kai Chen, Greg~S Corrado, and Jeff Dean. 2013.
\newblock Distributed representations of words and phrases and their
  compositionality.
\newblock In \emph{Advances in neural information processing systems}, pages
  3111--3119.

\bibitem[{Mishra et~al.(2020)Mishra, Arunkumar, Sachdeva, Bryan, and
  Baral}]{mishra2020dqi}
Swaroop Mishra, Anjana Arunkumar, Bhavdeep Sachdeva, Chris Bryan, and Chitta
  Baral. 2020.
\newblock Dqi: A guide to benchmark evaluation.
\newblock \emph{arXiv preprint arXiv:2008.03964}.

\bibitem[{Mishra et~al.(2022{\natexlab{a}})Mishra, Khashabi, Baral, Choi, and
  Hajishirzi}]{mishra-etal-2022-reframing}
Swaroop Mishra, Daniel Khashabi, Chitta Baral, Yejin Choi, and Hannaneh
  Hajishirzi. 2022{\natexlab{a}}.
\newblock \href {https://doi.org/10.18653/v1/2022.findings-acl.50} {Reframing
  instructional prompts to {GPT}k{'}s language}.
\newblock In \emph{Findings of the Association for Computational Linguistics:
  ACL 2022}, pages 589--612, Dublin, Ireland. Association for Computational
  Linguistics.

\bibitem[{Mishra et~al.(2022{\natexlab{b}})Mishra, Khashabi, Baral, and
  Hajishirzi}]{mishra2022cross}
Swaroop Mishra, Daniel Khashabi, Chitta Baral, and Hannaneh Hajishirzi.
  2022{\natexlab{b}}.
\newblock Cross-task generalization via natural language crowdsourcing
  instructions.
\newblock In \emph{Proceedings of the 60th Annual Meeting of the Association
  for Computational Linguistics (Volume 1: Long Papers)}, pages 3470--3487.

\bibitem[{Mishra and Sachdeva(2020)}]{mishra2020we}
Swaroop Mishra and Bhavdeep~Singh Sachdeva. 2020.
\newblock Do we need to create big datasets to learn a task?
\newblock In \emph{Proceedings of SustaiNLP: Workshop on Simple and Efficient
  Natural Language Processing}, pages 169--173.

\bibitem[{Ouyang et~al.(2022)Ouyang, Wu, Jiang, Almeida, Wainwright, Mishkin,
  Zhang, Agarwal, Slama, Ray et~al.}]{ouyang2022training}
Long Ouyang, Jeff Wu, Xu~Jiang, Diogo Almeida, Carroll~L Wainwright, Pamela
  Mishkin, Chong Zhang, Sandhini Agarwal, Katarina Slama, Alex Ray, et~al.
  2022.
\newblock Training language models to follow instructions with human feedback.
\newblock \emph{Preprint}.

\bibitem[{Parmar et~al.(2022)Parmar, Mishra, Purohit, Luo, Murad, and
  Baral}]{parmar2022boxbart}
Mihir Parmar, Swaroop Mishra, Mirali Purohit, Man Luo, M~Hassan Murad, and
  Chitta Baral. 2022.
\newblock In-boxbart: Get instructions into biomedical multi-task learning.
\newblock \emph{arXiv preprint arXiv:2204.07600}.

\bibitem[{Pennington et~al.(2014)Pennington, Socher, and
  Manning}]{pennington2014glove}
Jeffrey Pennington, Richard Socher, and Christopher~D Manning. 2014.
\newblock Glove: Global vectors for word representation.
\newblock In \emph{Proceedings of the 2014 conference on empirical methods in
  natural language processing (EMNLP)}, pages 1532--1543.

\bibitem[{Quionero-Candela et~al.(2009)Quionero-Candela, Sugiyama,
  Schwaighofer, and Lawrence}]{quionero2009dataset}
Joaquin Quionero-Candela, Masashi Sugiyama, Anton Schwaighofer, and Neil~D
  Lawrence. 2009.
\newblock \emph{Dataset shift in machine learning}.
\newblock The MIT Press.

\bibitem[{Sakaguchi et~al.(2019)Sakaguchi, Bras, Bhagavatula, and
  Choi}]{sakaguchi2019winogrande}
Keisuke Sakaguchi, Ronan~Le Bras, Chandra Bhagavatula, and Yejin Choi. 2019.
\newblock Winogrande: An adversarial winograd schema challenge at scale.
\newblock \emph{arXiv preprint arXiv:1907.10641}.

\bibitem[{Sanh et~al.(2021)Sanh, Webson, Raffel, Bach, Sutawika, Alyafeai,
  Chaffin, Stiegler, Scao, Raja et~al.}]{sanh2021multitask}
Victor Sanh, Albert Webson, Colin Raffel, Stephen~H Bach, Lintang Sutawika,
  Zaid Alyafeai, Antoine Chaffin, Arnaud Stiegler, Teven~Le Scao, Arun Raja,
  et~al. 2021.
\newblock Multitask prompted training enables zero-shot task generalization.
\newblock \emph{arXiv preprint arXiv:2110.08207}.

\bibitem[{Socher et~al.(2013)Socher, Perelygin, Wu, Chuang, Manning, Ng, and
  Potts}]{socher2013recursive}
Richard Socher, Alex Perelygin, Jean Wu, Jason Chuang, Christopher~D Manning,
  Andrew~Y Ng, and Christopher Potts. 2013.
\newblock Recursive deep models for semantic compositionality over a sentiment
  treebank.
\newblock In \emph{Proceedings of the 2013 conference on empirical methods in
  natural language processing}, pages 1631--1642.

\bibitem[{Swayamdipta et~al.(2020)Swayamdipta, Schwartz, Lourie, Wang,
  Hajishirzi, Smith, and Choi}]{swayamdipta2020dataset}
Swabha Swayamdipta, Roy Schwartz, Nicholas Lourie, Yizhong Wang, Hannaneh
  Hajishirzi, Noah~A Smith, and Yejin Choi. 2020.
\newblock Dataset cartography: Mapping and diagnosing datasets with training
  dynamics.
\newblock \emph{arXiv preprint arXiv:2009.10795}.

\bibitem[{Talmor and Berant(2019)}]{talmor2019multiqa}
Alon Talmor and Jonathan Berant. 2019.
\newblock Multiqa: An empirical investigation of generalization and transfer in
  reading comprehension.
\newblock \emph{arXiv preprint arXiv:1905.13453}.

\bibitem[{Varshney et~al.(2020)Varshney, Mishra, and Baral}]{varshney2020s}
Neeraj Varshney, Swaroop Mishra, and Chitta Baral. 2020.
\newblock It's better to say" i can't answer" than answering incorrectly:
  Towards safety critical nlp systems.
\newblock \emph{arXiv preprint arXiv:2008.09371}.

\bibitem[{Varshney et~al.(2022)Varshney, Mishra, and
  Baral}]{varshney2022investigating}
Neeraj Varshney, Swaroop Mishra, and Chitta Baral. 2022.
\newblock Investigating selective prediction approaches across several tasks in
  iid, ood, and adversarial settings.
\newblock In \emph{Findings of the Association for Computational Linguistics:
  ACL 2022}, pages 1995--2002.

\bibitem[{Wang et~al.(2018)Wang, Singh, Michael, Hill, Levy, and
  Bowman}]{wang2018glue}
Alex Wang, Amanpreet Singh, Julian Michael, Felix Hill, Omer Levy, and Samuel~R
  Bowman. 2018.
\newblock Glue: A multi-task benchmark and analysis platform for natural
  language understanding.
\newblock \emph{arXiv preprint arXiv:1804.07461}.

\bibitem[{Wei et~al.(2021)Wei, Bosma, Zhao, Guu, Yu, Lester, Du, Dai, and
  Le}]{wei2021finetuned}
Jason Wei, Maarten Bosma, Vincent~Y Zhao, Kelvin Guu, Adams~Wei Yu, Brian
  Lester, Nan Du, Andrew~M Dai, and Quoc~V Le. 2021.
\newblock Finetuned language models are zero-shot learners.
\newblock \emph{arXiv preprint arXiv:2109.01652}.

\bibitem[{Wieting et~al.(2015)Wieting, Bansal, Gimpel, and
  Livescu}]{wieting2015towards}
John Wieting, Mohit Bansal, Kevin Gimpel, and Karen Livescu. 2015.
\newblock Towards universal paraphrastic sentence embeddings.
\newblock \emph{arXiv preprint arXiv:1511.08198}.

\end{thebibliography}
\bibliographystyle{acl_natbib}

\clearpage
\appendix

\section{Analysis}\label{sec:appendix}

\begin{figure*}[]
    \centering
    \includegraphics[width=0.7\textwidth]{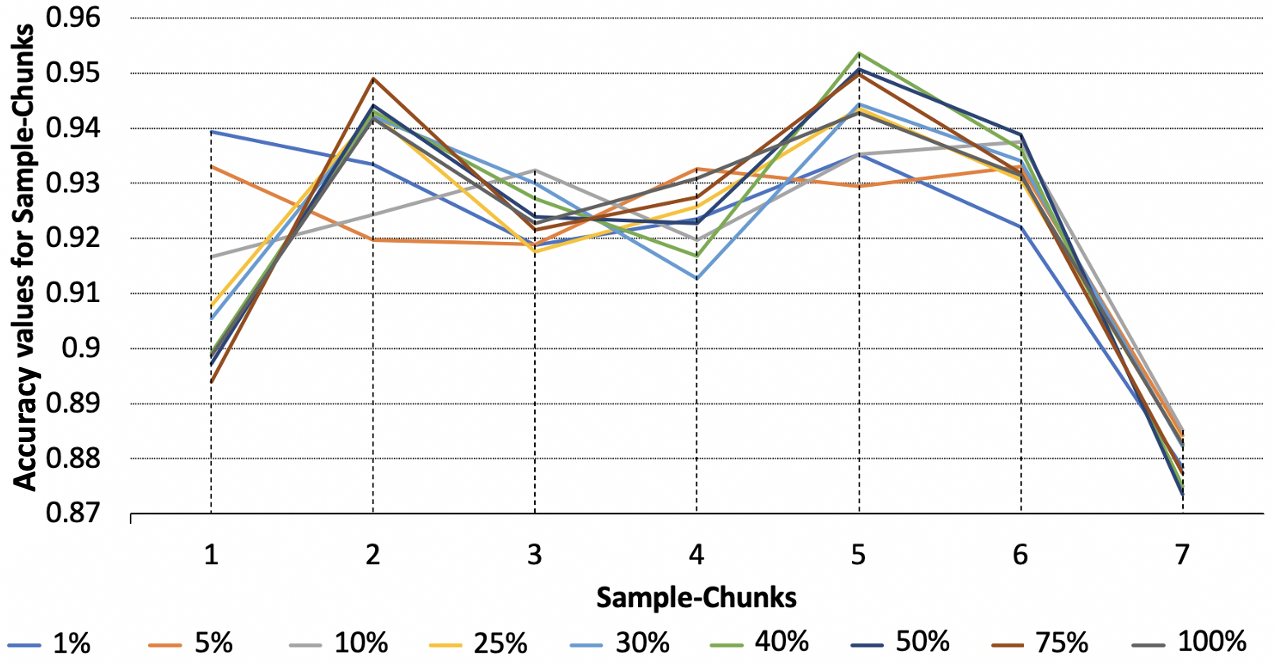}
    \caption{The top $b$\% of training samples is obtained by sorting in descending order of STS with each test set sample; test set samples are then divided into seven splits, based on decreasing STS averaged over the top $b$\% of training samples considered, for BERT-BASE over the SST-2 dataset.}
    \label{fig:5}
\end{figure*}

\begin{figure}[H]
    \centering
    \includegraphics[width=0.48\textwidth]{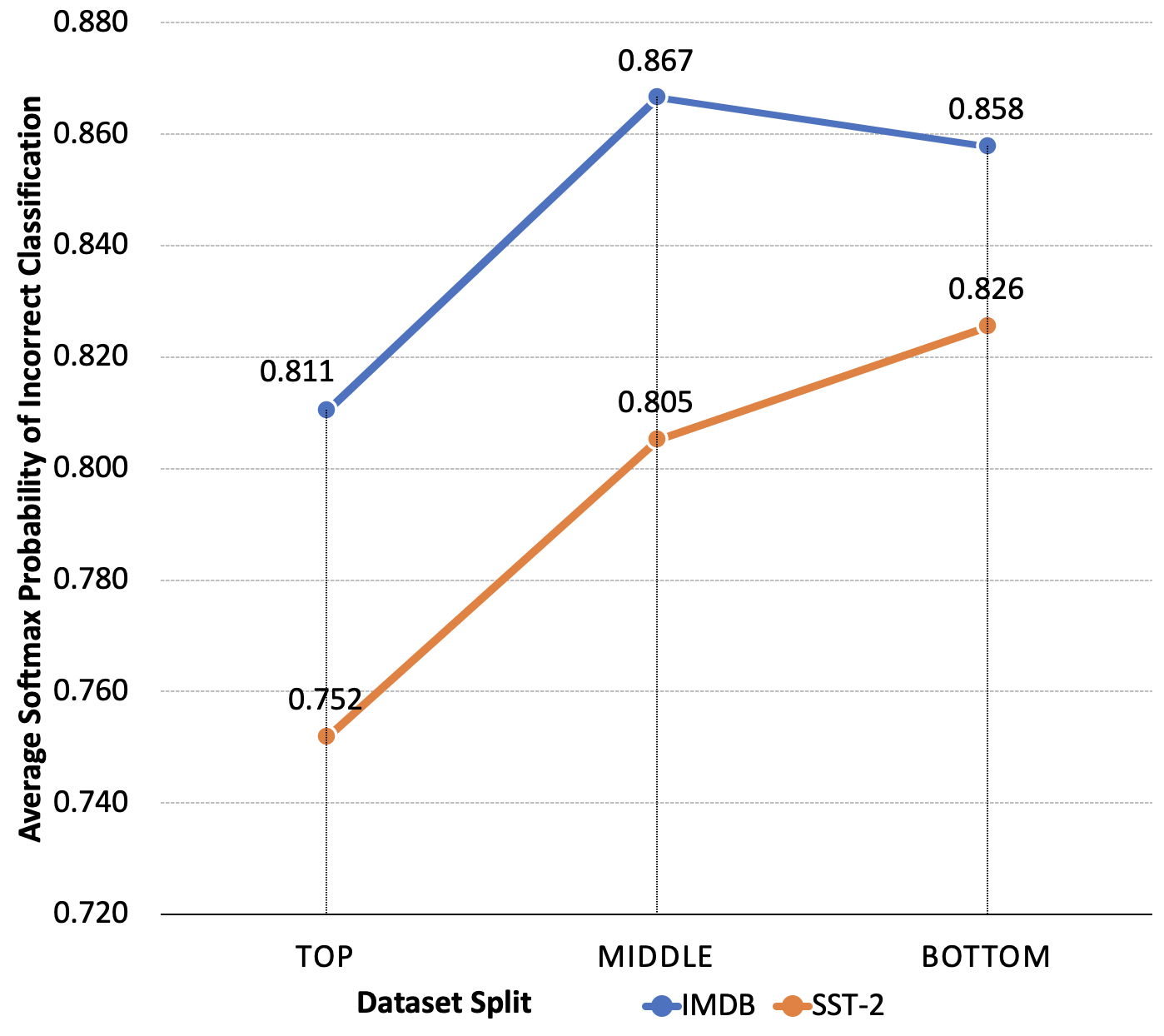}
    \caption{Average MaxProb of correct classifications using BERT-Base model across test sample-chunks of SST-2 and IMDB in decreasing order of train (IMDB)-test similarity. Monotonic slope decrease is desirable.}
    \label{fig:3}
\end{figure}
\begin{figure}[H]
    \centering
    \includegraphics[width=0.48\textwidth]{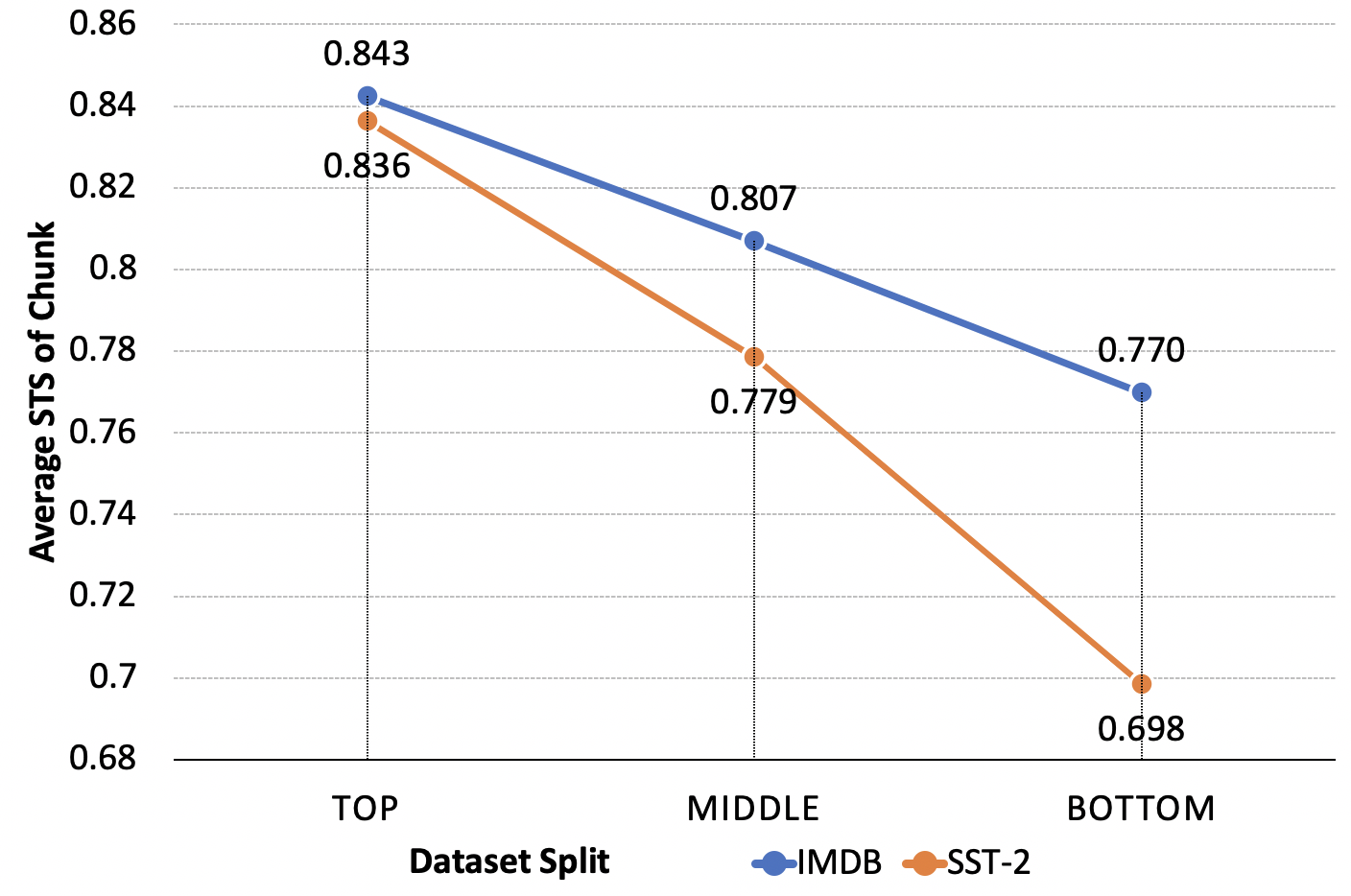}
    \caption{Average STS across test sample-chunks of SST-2 and IMDB in decreasing order of train (IMDB)-test similarity. Monotonic slope decrease is desirable.}
    \label{fig:comparests}
\end{figure}

\begin{figure*}[t]
    \centering
    \includegraphics[width=2\columnwidth]{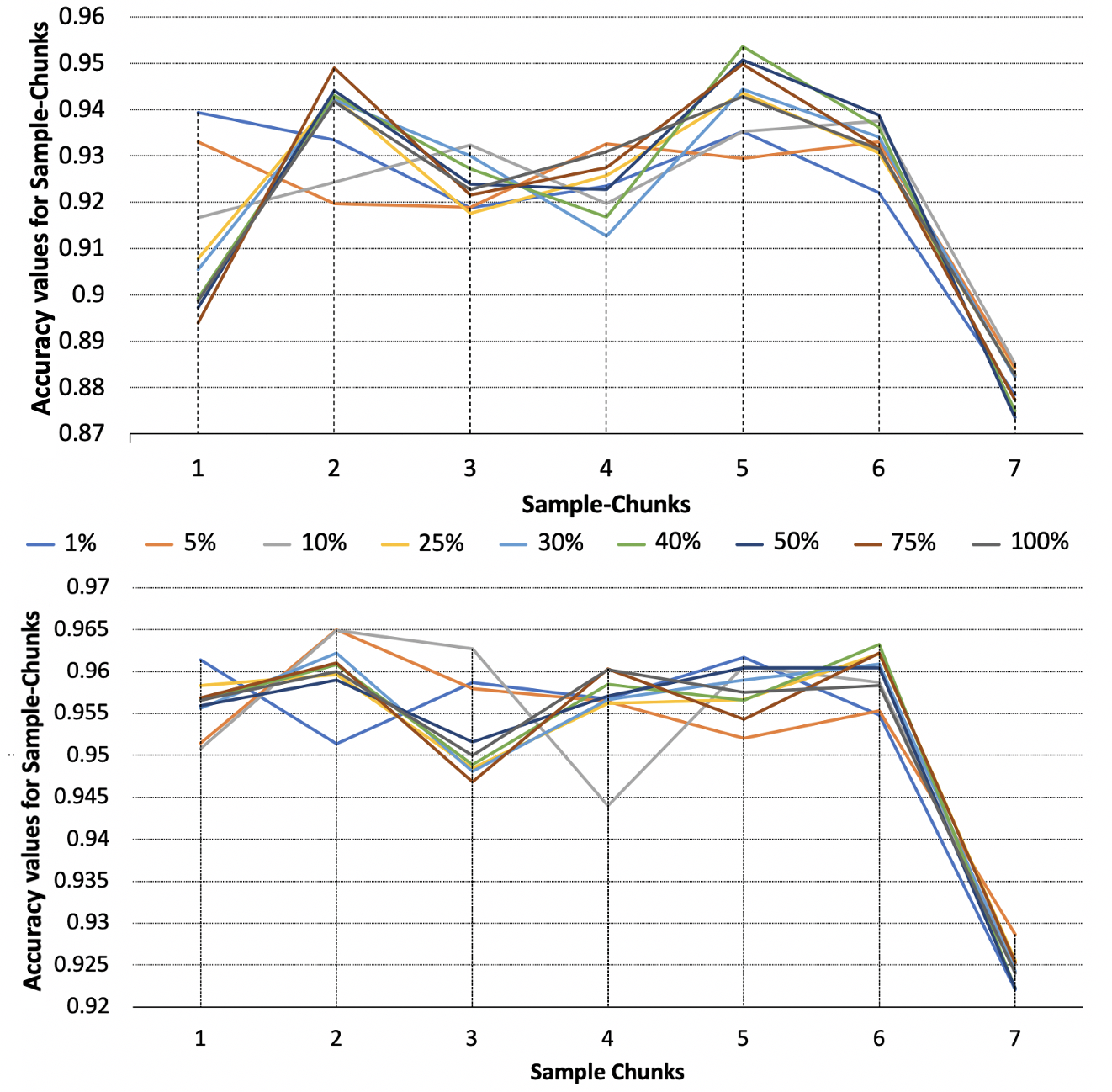}
    \caption{The top $b$\% of training samples is obtained by sorting in descending order of STS with each test set sample; test set samples are then divided into seven splits, based on decreasing STS averaged over the top $b$\% of training samples considered, for BERT-BASE over the SST-2 (top) and IMDB (bottom) datasets.}
    \label{roberta}
\end{figure*}

\begin{figure*}[t]
    \centering
    \includegraphics[width=2\columnwidth]{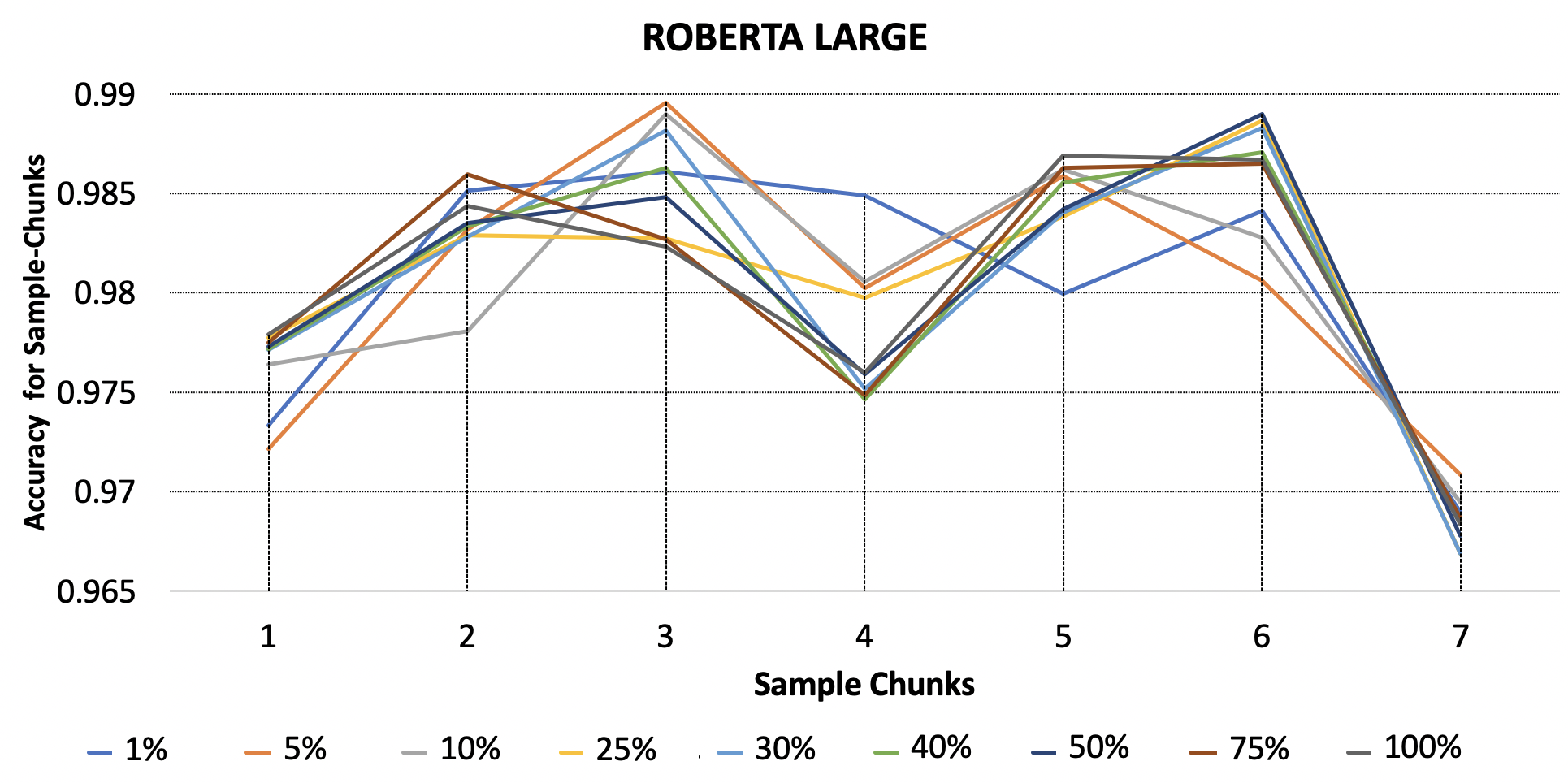}
    \includegraphics[width=2\columnwidth]{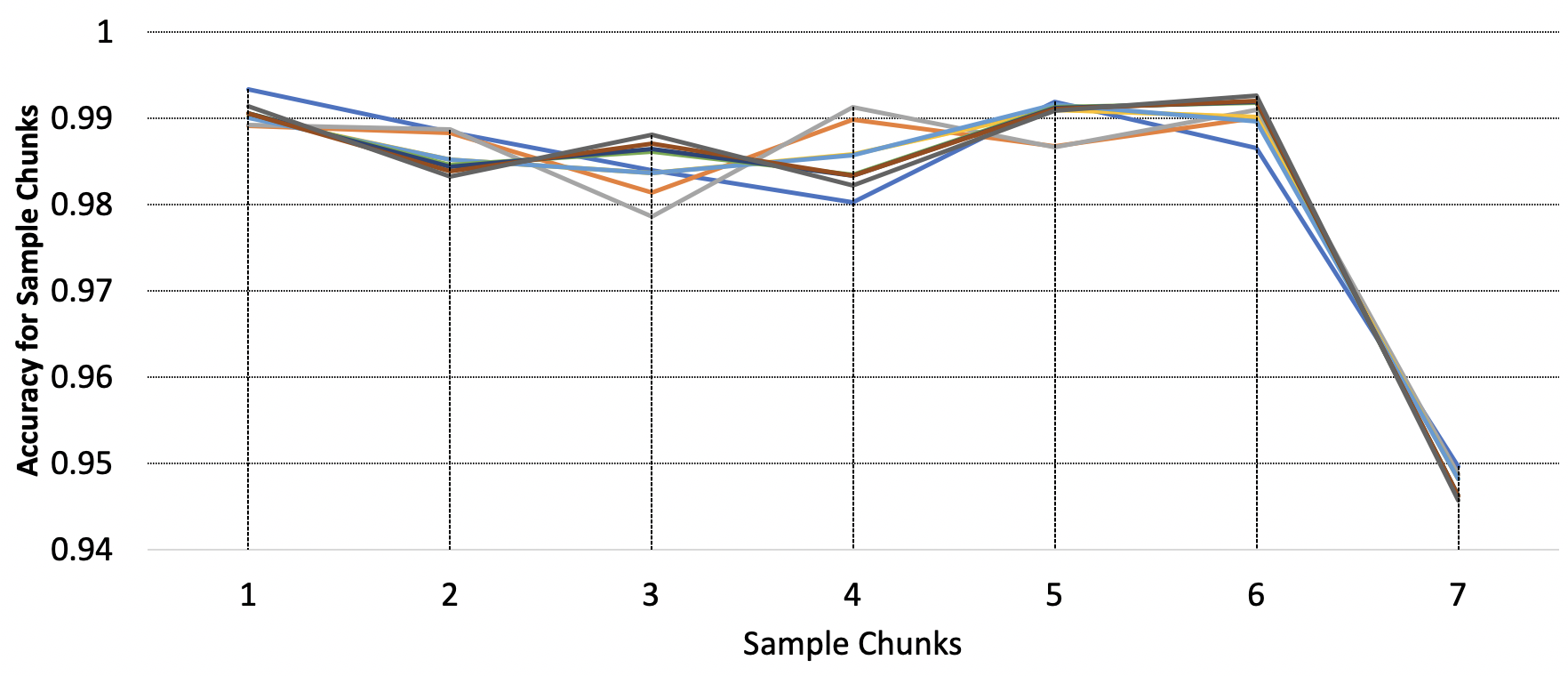}
    \caption{The top $b$\% of training samples is obtained by sorting in descending order of STS with each test set sample; test set samples are then divided into seven splits, based on decreasing STS averaged over the top $b$\% of training samples considered, for ROBERTA-LARGE over the SST-2 (top) and IMDB (bottom) datasets.}
    \label{roberta}
\end{figure*}

\begin{figure*}[t]
    \centering
    \includegraphics[width=2\columnwidth]{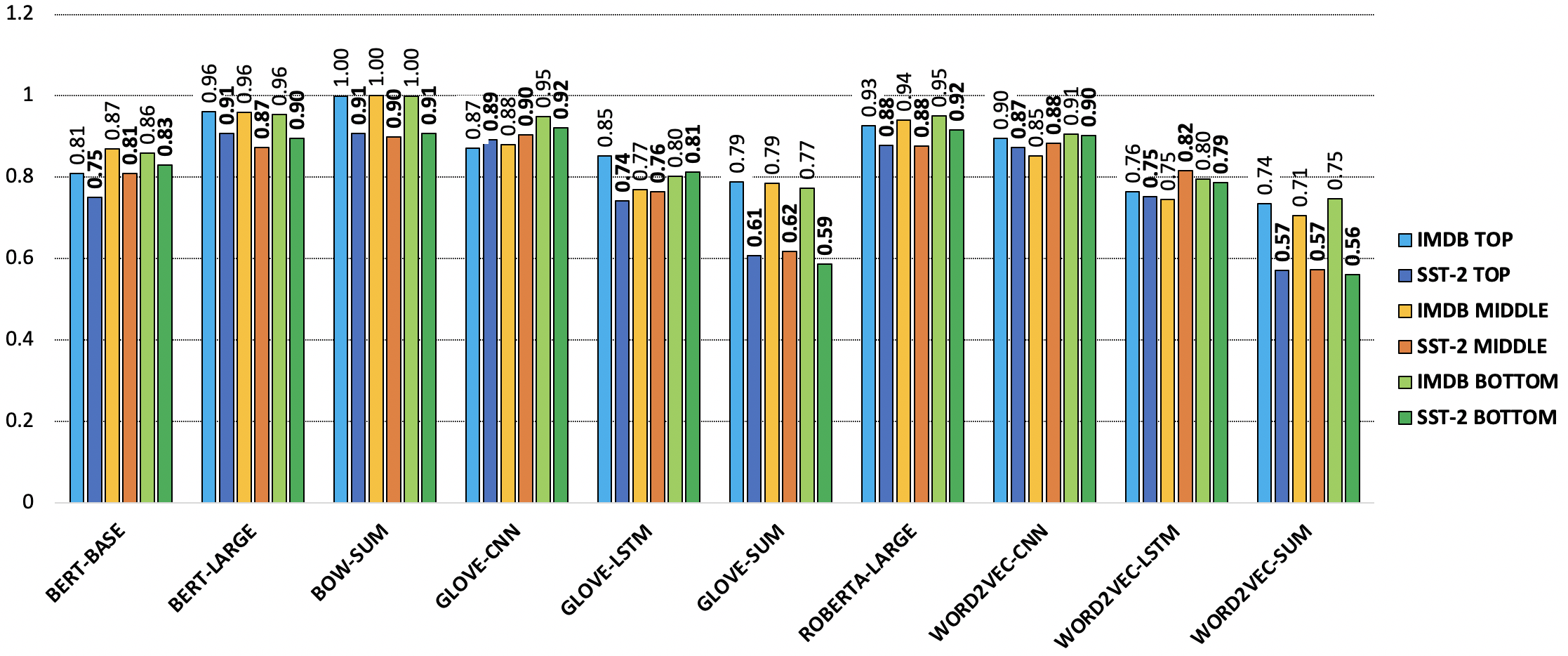}
    \caption{Confidence of incorrect classifications across test samples of SST-2 and IMDB in decreasing order of train (IMDB)-test similarity.}
\end{figure*}

\section{Infrastructure Used}
All the experiments were conducted on ”TeslaV100-SXM2-16GB”; CPU cores per node 20; CPU memory per node: 95,142 MB; CPU memory per core: 4,757 MB. This configuration is not a necessity for these experiments as we ran our operations with NVIDIA Quadro RTX 4000 as well with lesser memory. We used AllenNLP \cite{gardner2018allennlp} for our implementations.
\section{Another Case of STS Failure:}
Similar to Figure 6 of the main paper, we show a case for RoBERTA-Large where STS does not monotonically correlate with accuracy.
\section{STS's correlation with MaxProb Across Models}
Similar to our observation in Figure 2 of the main paper, here also we see that, in transformers, STS better correlates with MaxProb in comparison with other classes of models. This may further indicate the effectiveness of transformers in utilizing training data. Als, this observation opens up opportunity for further research as MaxProb has its own limitations and is not the ideal indicator of model confidence \cite{kamath2020selective, varshney2022investigating}.

\end{document}